\documentclass{article} 
\usepackage{nips13submit_e,times}
\usepackage{url}

\usepackage{graphicx} 
\usepackage{subfigure} 
\usepackage{multirow}
\usepackage{subfigure}
\usepackage{colortbl}

\usepackage{algorithm}
\usepackage{algorithmic}

\usepackage{amsmath}

\usepackage{color}
\usepackage{footmisc}

\usepackage{rotating}
\usepackage{multirow}

\usepackage{breakurl}
\usepackage[breaklinks]{hyperref}

\newcommand{\ourmethod}{SP-SVM}
\newcommand{\urlT}[1]{\url{#1}}

\newcommand{\x}{\mathbf{x}}
\newcommand{\set}[1]{\{#1\}}
\newcommand{\transpose}{\!^\top}

\newcommand{\J}{\mathcal{J}}

\newcommand{\K}{\mathbf{K}}
\newcommand{\bbeta}{\boldsymbol{\beta}}
\newcommand{\D}{\mathcal{D}}
\newcommand{\R}{\mathcal{R}}
\newcommand{\w}{\mathbf{w}}

\title{Parallel Support Vector Machines in Practice}

\author{
Stephen Tyree$^*$ \\
\texttt{swtyree@wustl.edu} \\
\And
Jacob R. Gardner$^*$ \\
\texttt{gardner.jake@wustl.edu} \\
\AND
Kilian Q. Weinberger$^*$ \\
\texttt{kilian@wustl.edu} \\
\And
Kunal Agrawal$^*$ \\
\texttt{kunal@wustl.edu} \\
\And
John Tran$^\dagger$ \\
\texttt{johntran@nvidia.com} \\
\AND
$^*$\textnormal{Washington University in St. Louis} \\
Department of Computer Science \& Engineering \\
St. Louis, MO, USA
\And
$^\dagger$\textnormal{NVIDIA} \\
St. Louis, MO, USA
}

\nipsfinalcopy 

\begin{document}

\maketitle

\begin{abstract}
In this paper, we evaluate the performance of various parallel
optimization methods for \textit{Kernel Support Vector Machines} on
multicore CPUs and GPUs. In particular, we provide the first
comparison of algorithms with \emph{explicit} and \emph{implicit}
parallelization. Most existing parallel implementations for multi-core
or GPU architectures are based on \emph{explicit} parallelization of
Sequential Minimal Optimization (SMO)---the programmers identified
parallelizable components and hand-parallelized them, specifically
tuned for a particular architecture. We compare these approaches with
each other and with \emph{implicitly parallelized} algorithms---where
the algorithm is expressed such that most of the work is done within few iterations with
large dense linear algebra operations. These can be computed with
highly-optimized libraries, that are carefully parallelized for a
large variety of parallel platforms. We highlight the advantages and
disadvantages of both approaches and compare them on various benchmark
data sets.We find an approximate implicitly parallel algorithm which is
surprisingly efficient, permits a much simpler implementation, and
leads to unprecedented speedups in SVM training.

\end{abstract}

\section{Introduction}\label{section:introduction}
Kernel support vector machines (SVM) are arguably among the most
established machine learning algorithms. They can capture complex,
nonlinear decision boundaries with good generalization to previously
unseen data.  Numerous specialized solvers
exist~\cite{chang2011libsvm,svmlight,smola2008bundle}, which take
advantage of the sparseness inherent in the optimization, and are
known to be effective on a large variety of classification problems.

Recently, trends in computer architecture have been moving toward
increasingly parallel hardware.  Most CPUs feature multiple cores, and
general purpose graphics processing units (GPUs) can execute thousands
of parallel threads on their hundreds of throughput-optimized cores.
Both parallel frameworks offer enormous raw power, and have the
potential to provide huge speedups; however, to utilize each type of
parallel thread effectively, algorithms must be carefully decomposed
and optimized in fundamentally different ways. For example, GPUs are
based on a ``same instruction multiple data'' (SIMD) architecture,
which requires all threads within one block to execute the exact same
instructions, whereas multi-core CPUs have much fewer threads with no
such restriction.  

On a high level, there are two different approaches to parallelizing
algorithms: \emph{Explicit} and \emph{implicit} approaches. In the
explicit approach, an algorithm is parallelized by hand --- that is,
the programmer finds the independent components of the algorithm which
can be run in parallel and encodes this parallelism using some
appropriate explicitly parallel language or library such as OpenMP
(for multicores), MPI (for clusters), CUDA or OpenCL (for GPUs).  In
the implicit approach, the algorithm is expressed as a
series of operations which are known to be highly parallel and for
which highly optimized parallel libraries already exist for most
platforms.  Examples include libraries for dense linear algebra
operations --- such as PLASMA~\cite{agullo2009magma} and Intel's
MKL~\cite{mkl} for multicores; MAGMA~\cite{agullo2009magma}, Jacket~\cite{jacket}, and CuBLAS~\cite{cublas} for GPUs --- and PDE solvers such as PETSc~\cite{petsc-user-ref}.

Both approaches have advantages and disadvantages. The explicit
approach can be applied to most algorithms; therefore, in particular,
it can probably be applied to the exact algorithm of one's choice.  However,
it often requires a significant engineering effort and a fine-tuned
tradeoff between parallel work and induced overhead---which needs to
be calibrated specifically for any particular algorithm and parallel
architecture.  The implicit approach is only applicable if the
algorithm in question can be formulated as operations of some
well-optimized library (in our case, linear algebra operations), which
may not always be possible or may require approximation or
relaxation of the problem, potentially leading to a loss in accuracy.  If it is
possible, however, the implicit approach has two advantages. First, since
researchers and engineers have carefully designed and optimized these
linear algebra libraries for peak
performance~\cite{agullo2009magma,cublas}, they typically provide
great speedups as long as they are called on sufficiently large
problems.  Therefore, if we can express an algorithm in terms of linear
algebra operations of large-enough granularity, implicit algorithms
can provide great parallel speedups, often more so than explicit
algorithms.  Second, these libraries are maintained and ported to new hardware as it
becomes available; therefore, there is no need to rewrite an implicit
algorithm for each new generation. In light of these two options,  we investigate the following question: \emph{Given recent changes in hardware design, which approach to kernel SVM parallelization is most efficient?}

To our knowledge, all existing (competitive) parallel SVM
implementations for multi-core or GPU
systems~\cite{athanasopoulos2011gpulibsvm, carpenter2009cusvm,
  catanzaro2008, chang2011libsvm, cotter2011gtsvm} use the explicit
parallelization approach on dual decomposition methods, such as
Platt's SMO algorithm~\cite{platt1998smo}.  Although implicit parallelization comes naturally
for \emph{e.g.} deep neural nets~\cite{krizhevsky2012imagenet}, it does not initially fit the SVM formulation and until this work there
were no comparable SVM implementations of implicit parallelization. 
However, there exist at least three publications that
reduce the kernel SVM optimization to dense linear algebra
operations. Sha et al.~\cite{sha2007mult} introduce a multiplicative
update rule for the exact SVM optimization problem, which uses large
matrix-vector multiplications in each
iteration. Chapelle~\cite{chapelle2007primal} proposes a primal
formulation for the least squares hinge loss~\cite{suykens1999least}
which results in matrix-matrix and matrix-vector operations, and
Keerthi et al.~\cite{keerthi2006newton} approximates this approach by
restricting the support vectors to a smaller subset (for reduced
test-time complexity).

One advantage of the implicitly parallel approach is that, if done
correctly, the algorithm spends almost all of its execution time in highly
optimized routines and very little time in the remainder of the
program, which therefore can be written in a high level language like
\verb!MATLAB! or \verb!Python!. This enabled us to implement implicit
parallel versions of all three approaches, which naturally work on both multi-core and GPU systems, by linking against appropriate algebra
libraries~\cite{mkl,cublas}.

We apply an empirical approach and compare the various implementations
with each other on several medium-sized classification data sets on
GPU and multi-core architectures and arrive at an interesting
conclusion: Although the multiplicative update rule~\cite{sha2007mult}
and the primal optimization~\cite{chapelle2007primal} do not scale to
our data set sizes due to their quadratic memory complexity,
Keerthi's~\cite{keerthi2006} sparse primal optimization appears to be
an excellent compromise. Our \verb!MATLAB! implementation tends to consistently
outperform all highly optimized explicitly parallel algorithms and
generally suffers no or little decrease in accuracy
due to the problem relaxations.

In this paper we make two contributions: 1. We provide the first
detailed empirical analysis of both explicit and implicit SVM
parallelization for multi-core CPUs and GPU architectures; 2. We observe that
implicit parallelization can be a much more efficient approach where possible.
We believe that these insights are valuable to the machine
learning community, which has so far focused almost entirely on
explicit parallelism, and encourage further research into implicit
approaches to parallelism.

\section{Notation and Background}
\label{section:svm}

Throughout this paper we type vectors in bold ($\x_i$), scalars in
regular ($C$ or $b$), matrices in capital bold ($\K$) and sets in
cursive ($\J$) font. Specific entries in vectors or matrices are
scalars and follow the corresponding convention, \emph{i.e.} the
$i^{th},j^{th}$ entry of matrix $\K$ is written as $K_{ij}$ and the
$i^{th}$ dimension of vector $\x$ is $x_i$. In contrast, depending on
the context, $\x_i$ refers to the $i^{th}$ vector within some ordered
set $\x_1,\dots,\x_n$ and $\mathbf{k}_i$ refers to the $i^{th}$ column
in a matrix $\K$.

\paragraph{Kernelized SVMs.}  When training a support vector machine, we are given a training dataset $\D = \set{(\x_1,y_1),...,(\x_n,y_n)}$ of feature vectors $\x_i \in \R^d$ with class labels $y_i \in \set{-1,+1}$. The goal of the optimization is to find a maximum margin hyperplane separating the two classes.
(Binary classifiers can easily be extended to multiclass settings through pairwise coupling or similar approaches~\cite{scholkopf2001learning}.) 
The primal formulation of the SVM optimization problem \cite{Cortes95svms} learns a hyperplane parameterized by weight vector $\w$ with a scalar offset $b$:
\begin{equation}\label{eq:primal}
    \underset{\w,b}{\min} \frac{1}{2} ||\w||^2 + C \sum_{i=1}^{n} \max(0, 1 - y_i (\w\transpose \x_i + b)).
\end{equation}
The simple linear case can be solved very efficiently with special purpose algorithms~\cite{LIBLINEAR}. In this paper we focus on non-linear SVMs, which map the inputs  into a new feature space $\x_i \to \phi(\x_i)$ prior to optimizing, where $\phi(\x_i)$ is a nonlinear transformation of $\x_i$. This mapping is generally to a higher (possibly infinite) dimensional representation. 
As inputs are only accessed through pairwise inner products in the dual formulation of the optimization, the mapping can be computed implicitly with the  \emph{kernel-trick}~\cite{scholkopf2001learning} through a positive semi-definite kernel function $k(\x_{i},\x_{j}) \!=\! \phi(\x_i)\transpose\phi(\x_j)$. The (dual) optimization to find the large-margin hyperplane becomes
\begin{equation}\label{eq:dualkernel}
\underset{C\geq \alpha_{i}\geq0}{\max}-\frac{1}{2}\sum_{i=1}^{n}\sum_{j=1}^{n}\alpha_{i}\alpha_{j}y_{i}y_{j}k(\x_{i},\x_{j})+\sum_{i=1}^{n}\alpha_{i},
\end{equation}
where a Lagrange multiplier variable $\alpha_{i}$ corresponds to each training input. At the end of the optimization, only some variables $\alpha_{i}$ are nonzero, which are referred to as \emph{support vectors}. 
(For convenience, henceforth, we omit the bias term $b$, which can be solved for in a straight-forward fashion from the solution of (\ref{eq:dualkernel})~\cite{scholkopf2001learning}.
Throughout this paper we will focus primarily on the Radial Basis Function (RBF) kernel: $k(\x_{i},\x_{j}) = e^{-\gamma ||\x_i - \x_j||^{2}}$. The RBF kernel is particularly interesting because of its universal approximation properties~\cite{scholkopf2001learning} and its wide-spread application. 

Although solving the SVM optimization in the dual formulation (\ref{eq:dualkernel}) avoids the explicit feature computation $\phi(\x_i)$, it is still significantly slower than solving the linear formulation. 
In particular, it requires either precomputing the kernel matrix $\K$ where $K_{ij}=k(\x_i,\x_j)$, requiring $O(n^{2})$ space, or recomputing $k(\x_i,\x_j)$ as it is needed, with space or time complexity that is too great for ever increasing data set sizes. This motivates the adoption of SVM-specific optimization procedures.


\section{Explicitly Parallel SVM Optimization}
\label{section:explicit}

To our knowledge, all competitive implementations of parallel SVMs
(for multi-core CPUs or GPU architectures) are based on explicit
parallelization of dual decomposition approaches.  Dual decomposition
methods, which include Sequential Minimal Optimization (SMO)
\cite{platt1998smo}, are among the most efficient sequential
algorithms for solving the dual formulation.  They operate on a small
\emph{working set} of Lagrange multiplier variables in each iteration,
holding others constant.  For example, in each iteration, SMO
heuristically selects two dual variables $\alpha_i,\alpha_j$ and
optimizes them analytically.  LibSVM, a very popular tool for training
SVMs, implements a variant of this method \cite{chang2011libsvm}.  In
general, any small number of dual variables may be optimized at once
with working set size representing a tradeoff between work per
iteration and number of iterations required.  Explicit parallelization
approaches parallelize the computation within
each iteration as well as parallelizing kernel computations.  
A common theme among explicitly parallel methods is high code complexity,
making it hard to verify correctness or port the code to new or updated hardware platforms.

\paragraph{Multi-core.}
There are several parallel implementations of dual decomposition-based
SVM solvers targeted toward multi-cores.  Some methods attempt to
extract existing parallelism from SMO-based approaches
\cite{dong2003parallel,eitrich2006data}, including a simple
modification to LibSVM that computes kernel matrix entries in
parallel with OpenMP.  Other approaches attempt some restructuring of
the problem.  Increasing the working set size (originally two
variables in SMO) exposes additional parallelism, as several dual
variables are optimized at each iteration
\cite{brugger2007parallel,eitrich2005efficient,zanni2006parallel}, as
does optimizing over nested working sets \cite{zhao2011parallel}.
Another common approach is to partition the training set, optimize
over the partitions in parallel, and combine the resulting solutions
\cite{cao2006parallel,collobert2002parallel,graf2004parallel,hazan2008parallel,zhang2005cpu}.
We were only able to obtain source code for two of these methods ---
namely LibSVM with OpenMP and PSVM\cite{NIPS2007_435}.  We only report the results of the former, since the
latter was not designed for multi-core CPUs and consumed an infeasible amount
of memory for medium-scale datasets. However, a comparison of published training times (with
consideration of the various architectures) makes us believe that most
other approaches are comparable or (more often) less competitive in
practice.

\paragraph{GPU.}
Likewise, all previous attempts to accelerate the training of kernelized SVMs on
GPUs have been direct implementations of a dual decomposition method
such as SMO.  
GPU SVM \cite{catanzaro2008} offloads computation of kernel matrix rows to the GPU using the CUBLAS library and computes KKT condition updates on the GPU with explicitly parallelized routines.  A similar approach and results were demonstrated
by \cite{carpenter2009cusvm}.  
GTSVM \cite{cotter2011gtsvm} takes the strategy of increasing the
working set size of dual variables to $16$ to better utilize GPU
resources.  The method features built-in support for both multi-class
SVMs and sparse training vectors.  GTSVM achieves the best previously
published kernel SVM training times of which we are aware.  Other GPU
implementations include solvers especially optimized for multi-class
problems \cite{multisvm} and a specialized implementation in
\verb!R!~\cite{rpusvm}.

\section{Implicitly Parallel SVM Optimization}
\label{section:implicit}

As an alternative to explicitly parallelized SMO-type
optimization methods, we also investigate algorithms that are
amenable to implicit parallelization; that is, algorithms where
the majority of the work can be expressed in few iterations with dense linear algebra
computations, which can then be performed using optimized
libraries. 
We identify three re-formulations of the SVM problem
that lend themselves towards this approach, while noting that
none of these methods were explicitly developed for increased
parallelism. It is important to point out that in all formulations in this section, the linear algebra computations are dense irrespective of the sparsity of the data, as they operate on the dense kernel matrix, \emph{e.g.} computing Hessian updates. 
One downside of this implicit approach is that it sometimes requires a
reformulation or relaxation of the SVM optimization in
(\ref{eq:dualkernel}), which can impact accuracy and memory
efficiency.

\paragraph{Multiplicative update.}
Sha et al.\cite{sha2007mult} proposed the multiplicative update rule,
which updates all dual variables $\alpha_i$ in each iteration, to
solve the dual optimization (\ref{eq:dualkernel}).  This approach
relies on matrix-vector multiplication which can be readily
parallelized; the authors remark in their original
publication that the algorithm could potentially be used for parallel
implementations.  While our implementation demonstrated some speedups
when linked against parallel libraries, the method was ultimately
considered not competitive (and is not included in our experimental
section) for two reasons: 1. The entire kernel matrix must be stored
in memory at all times, which renders the method infeasible for
typical medium-sized data sets; and 2. the convergence rate of the
multiplicative update is too slow in practice, requiring too many iterations.

\paragraph{Primal optimization.}
Chapelle introduced a method for solving a kernel SVM optimization
problem  in the primal~\cite{chapelle2007primal}.  
The SVM classifier can be expressed as $h(\x)\!=\!\w^\top\phi(\x)+b$, where $\w\!=\! \sum_{i=1}^{n} \alpha_i y_i \phi(\x_i)$ (and with bias $b$). 
After the transformation $\x\rightarrow \phi(\x)$, solving (\ref{eq:primal}) with respect to $\w$ directly is impractical, due to the high (possibly infinite) dimensionality of $\phi(\x)$. However, after a change of variable, with $\beta_i\!=\!\alpha_iy_i$ and  $\bbeta \in \R^n$, (\ref{eq:primal}) can be rewritten as follows:
\begin{equation}\label{eq:primalK}
\min_{\bbeta,b} \frac{1}{2} \bbeta\transpose \K \bbeta + \frac{C}{2} \sum_{i=1}^{n} \max(0,1 - y_i (\bbeta\transpose \mathbf{k}_{i}+b)]^2
\end{equation}
where $\mathbf{k}_{i}$ is the kernel matrix row corresponding to the
$i^{th}$ training example. Notice that there are two relaxations:
1. the $\beta_i$ are unconstrained, in contrast to $\alpha_i$ in
(\ref{eq:dualkernel}), which must satisfy $0\!\leq \!\alpha_i\!\leq\!
C$; and 2.  the squared hinge loss is used in place of the more common
absolute hinge loss. These changes allow the use of second order
optimization methods. In particular Newton's method yields very fast
convergence with computations expressed as dense linear algebra
operations. As noted in \cite{chapelle2007primal}, the squared hinge
loss leads to almost identical results as the absolute hinge loss---a
claim that we confirm in our experimental results. Similar to the
multiplicative approach, this method requires the computation of the
entire kernel matrix, which renders it impractical for larger data
sets. We therefore do not include it in our experimental result
section, which focuses on data sets with prohibitively large sizes.

\paragraph{Sparse primal optimization.}
Keerthi et al. proposed a method to reduce the complexity of
Chapelle's primal approach by restricting the support vectors to some
subset of \emph{basis vectors} $\J \subset \{1,\dots,n\}$ so that $j
\notin \J \Rightarrow \beta_j=0$. Then equation (\ref{eq:primalK})
becomes:
\begin{equation}\label{eq:sparseobj}
	\min_{\bbeta,b} \frac{1}{2} \bbeta\transpose \K_{\J\J} \bbeta + \frac{C}{2} \sum_{i=1}^{n} \max(0, 1 - y_i (\bbeta\transpose \mathbf{k}_{\J i}+b))^2.
\end{equation}
Here, $\bbeta$ has been restricted to contain only those $\beta_i$
with $i\in\J$. $\K_{\J\J}$ is the kernel matrix between only basis
vectors, and $\mathbf{k}_{\J i}$ is the kernel row of the $i^{th}$
training example with all basis vectors (i.e., the vector
$\mathbf{k}(\x_k,\x_i)$ for each $k \in \J$).  As the set $\J$ is
originally unknown, Keerthi et al. propose to grow $\J$ with a
heuristic. Initially, $\J$ is empty and the algorithm then has two
distinct stages that are cycled.  \emph{Basis vector selection:} A
small subset of the training set is randomly sampled, and then a
heuristic is used to estimate the reduction in loss from adding each
input to $\J$. The highest scoring point is then greedily added to
$\J$ to get $\J^{'}$.  \emph{Reoptimization:} After a certain number
of basis vectors have been added to $\J^{'}$, (\ref{eq:sparseobj}) is
optimized using $\J^{'}$ as the basis vector set.
This whole process of gradually selecting basis vectors and then
re-optimizing repeats until some stopping criterion is met.  The
resulting algorithm performs only a few iterations in total, each of which
make use of intensive linear algebra computation.  This method still
requires the kernel matrix of basis vectors with all training
examples, requiring $O(|\J|n)$ space. In practice, $|\J| \ll n$;
however, this may still be a concern, particularly on GPUs where memory availability is more limited than RAM.

We reimplement this sparse primal SVM (SP-SVM) in
\verb!MATLAB!. For linear algebra operations on multicores, we
use a combination of built-in linear algebra functions and Intel
MKL. For linear algebra operations on the GPU, we use
Jacket\cite{jacket}, a \verb!MATLAB! toolkit for accelerating
computations on GPUs. Additionally, we incorporate the freely
available C++/CUDA package CUBLAS \cite{cublas} in cases where
Jacket proves to be inefficient or lacks desired functionality.
Because no stopping criterion is suggested in the original
publication~\cite{keerthi2006}, our implementation stops when,
after re-optimization, the change in training error divided by
the number of basis vectors added in the previous selection stage
is less than some threshold $\epsilon$. We have released an
optimized C++ version of SP-SVM, called WU-SVM, for both multicore
and GPU architectures at \mbox{\url{http://tinyurl.com/wu-svm}}.

\section{Results}\label{section:experiments}
%
\definecolor{gray1}{gray}{0.85}
\definecolor{gray2}{gray}{0.70}
\definecolor{blue1}{rgb}{0.6,0.8,1.0}
\definecolor{blue2}{rgb}{0.4,0.6,1.0}
\definecolor{red1}{rgb}{0.8,0.2,0.2}
\definecolor{red2}{rgb}{1.0,0.5,0.5}

\newcommand{\Cs}{&}
\newcommand{\Cm}{&\cellcolor{blue1}}
\newcommand{\Cg}{&\cellcolor{blue2}}
\newcommand{\CR}{&\cellcolor{red1}}
\newcommand{\Cr}{&\cellcolor{red2}}
\newcommand{\std}[1]{}
\newcommand{\na}{$-$}
\newcommand{\bad}{\red{!}}
\newcommand{\SPSVMMC}{SP-SVM}
\newcommand{\SPSVMGPU}{SP-SVM}
\newcommand{\LibSVMMC}{LibSVM}
\newcommand{\red}[1]{\textcolor{red1}{#1}}
\newcommand{\sclabel}{\hspace{-3pt}\multirow{1}{*}{\scriptsize\begin{sideways}SC\:\end{sideways}\hspace{-3pt}}}
\newcommand{\mclabel}{\hspace{-3pt}\multirow{-2}{*}{\scriptsize\begin{sideways}MC\;\end{sideways}\hspace{-3pt}}}
\newcommand{\gpulabel}{\hspace{-3pt}\multirow{-3}{*}{\scriptsize\begin{sideways}GPU\;\end{sideways}\hspace{-3pt}}}

\begin{table*}[p]
\begin{center}
\begin{tabular}{c|cc|c|c|c}
\textbf{Data Set}    &  \multicolumn{2}{c|}{\textbf{Method}}  & \textbf{Test Error (\%)}  & \textbf{Training Time}  & \textbf{Speedup} \\

\hline 
\em Adult  							\Cs	\sclabel	\Cs LibSVM  \Cs 14.9 \Cs 1m 6s   \Cs 1$\times$   \\ 
\small$7$\scriptsize{MB}			\Cm				\Cm \textbf{\LibSVMMC}   \Cm 14.9 \Cm \textbf{10.5s} \Cm \textbf{18$\times$}   \\
\small$n\!=\!31562,\ d\!=\!123$ 	\Cm	\mclabel	\Cm \SPSVMMC   \Cm 14.8\std{0.04}   \Cm 15.2s\std{5.2s}   \Cm 13$\times$   \\
\small$C\!=\!1, \gamma\!=\!0.05$ \cite{cotter2011gtsvm}
									\Cg				\Cg GPU SVM    \Cg 14.9 \Cg 6s  \Cg 32$\times$   \\
									\Cg				\Cg \textbf{GTSVM}	\Cg 14.8 \Cg \textbf{1s} \Cg \textbf{190}$\times$   \\
									\Cg	\gpulabel	\Cg \SPSVMGPU   \Cg 14.8\std{0.06}   \Cg 11.3s\std{3.0s}   \Cg 17$\times$   \\
 
\hline 
\em Covertype/Forest  				\Cs	\sclabel	\Cs LibSVM \Cs 13.9 \Cs 5h 1m 19s    \Cs 1$\times$   \\
\small$96$\scriptsize{MB} 			\Cm				\Cm \LibSVMMC    \Cm 13.9 \Cm 1h 5m 46s    \Cm 5$\times$   \\
\small$n\!=\!522911,\ d\!=\!54$ 	\Cm	 \mclabel	\Cm \textbf{\SPSVMMC}    \Cm 13.7\std{0.02}   \Cm \textbf{10m 10s\std{10s}}  \Cm \textbf{29$\times$}   \\
\small$C\!=\!3, \gamma\!=\!1$ \cite{cotter2011gtsvm}
									\Cg				\Cg GPU SVM \Cg 13.9 \Cg 7m 32s \Cg 40$\times$   \\
									\Cg				\Cg GTSVM  \Cg \red{36.8}    \Cg \red{5m 15s}    \Cg \red{57$\times$}   \\
									\Cg \gpulabel	\Cg \textbf{\SPSVMGPU}    \Cg 13.8\std{0.05}    \Cg \textbf{4m 38s}\std{30s}   \Cg \textbf{65}$\times$   \\

\hline
\em KDDCup99  						\Cs	\sclabel	\Cs LibSVM  \Cs 7.4 \Cs 3h 0m 29s    \Cs 1$\times$  \\
\small$970$\scriptsize{MB} 			\Cm				\Cm \LibSVMMC    \Cm 7.4 \Cm 26m 37s \Cm 7$\times$  \\
\small$n\!=\!4898431,\ d\!=\!127$  	\Cm	 \mclabel	\Cm \textbf{\SPSVMMC}    \Cm 7.9\std{0.23}   \Cm \textbf{56s\std{14.0s}}    \Cm \textbf{193$\times$}  \\
\small$C\!=\!10^6, \gamma\!=\!0.137$ \cite{tsang2006core}
									\Cg				\Cg  GPU SVM \Cg \na \Cg \na  \Cg \na \\
									\Cg 			\Cg  GTSVM \Cg \red{19.9}   \Cg \red{1h 15m 39s}  \Cg \red{2$\times$}  \\
									\Cg \gpulabel	\Cg  \SPSVMGPU    \Cg \na \Cg \na  \Cg \na \\
 

\hline
\em MITFaces  						\Cs	\sclabel	\Cs LibSVM \Cs 5.6$^\dagger$    \Cs 34m 22s    \Cs 1$\times$ \\
\small$1.3$\scriptsize{GB} 			\Cm				\Cm \textbf{\LibSVMMC}    \Cm 5.6$^\dagger$    \Cm \textbf{4m 8s}    \Cm \textbf{8$\times$} \\
\small$n\!=\!489410,\ d\!=\!361$  	\Cm	 \mclabel	\Cm \SPSVMMC   \Cm \red{7.4\std{?}$^\dagger$} \Cm \red{20s\std{0.9s}}  \Cm \red{103$\times$} \\
\small$C\!=\!20, \gamma\!=\!0.02$ \cite{tsang2006core}
									\Cg				\Cg \textbf{GPU SVM}    \Cg 5.7$^\dagger$   \Cg \textbf{33s}    \Cg \textbf{61$\times$}\\
									\Cg				\Cg GTSVM  \Cg 5.6$^\dagger$    \Cg 1m 34s    \Cg 22$\times$ \\
									\Cg \gpulabel	\Cg \SPSVMGPU   \Cg \red{7.4\std{?}$^\dagger$} \Cg \red{10s\std{2s}}   \Cg \red{200$\times$} \\
 
\hline 
\em FD  							\Cs	\sclabel	\Cs LibSVM \Cs 1.4  \Cs 2h 6m 50s   \Cs 1$\times$  \\   
\small$1.3$\scriptsize{GB}			\Cm				\Cm \LibSVMMC    \Cm 1.4   \Cm 27m 54s   \Cm 5$\times$   \\
\small$n\!=\!200000^*,\ d\!=900$  	\Cm	 \mclabel	\Cm \textbf{\SPSVMMC}    \Cm 1.5\std{0.03}    \Cm \textbf{1m 22s\std{4.0s}} \Cm \textbf{92$\times$}   \\
\small$C\!=\!10, \gamma\!=\!1$ 		\Cg				\Cg GPU SVM \Cg 1.4  \Cg 6m 20s    \Cg 20$\times$ \\
									\Cg				\Cg GTSVM  \Cg 1.5  \Cg 2m 26s    \Cg 52$\times$ \\
									\Cg \gpulabel	\Cg \textbf{\SPSVMGPU}    \Cg 1.5\std{0.02}    \Cg \textbf{29s}\std{0.6s}  \Cg \textbf{262}$\times$ \\

\hline    
\em Epsilon  						\Cs	\sclabel	\Cs LibSVM \Cs 10.9  \Cs 19h 12m 27s   \Cs 1$\times$ \\
\small$2.4$\scriptsize{GB} 			\Cm				\Cm \LibSVMMC    \Cm \na   \Cm \na \Cm \na \\
\small$n\!=\!160000^*,\ d\!=\!2000$	\Cm	 \mclabel	\Cm \textbf{\SPSVMMC}   \Cm 10.8\std{0.03}    \Cm \textbf{8m 10s\std{59s}}  \Cm \textbf{141$\times$} \\
\small$C\!=\!1, \gamma\!=\!0.125$	\Cg				\Cg  GPU SVM \Cg 10.9  \Cg 29m 1s    \Cg 40$\times$ \\
									\Cg 			\Cg  GTSVM \Cg 10.9  \Cg 4m 33s    \Cg 253$\times$ \\
									\Cg \gpulabel	\Cg \textbf{\SPSVMGPU} \Cg 10.8\std{0.06} \Cg \textbf{1m 55s}\std{12s} \Cg \textbf{601}$\times$ \\

\hline
\em MNIST8M \small($24$\scriptsize{GB}\small)
									\Cs	\sclabel	\Cs LibSVM \Cs 1.0   \Cs 12d 15h 21m 31s  \Cs 1$\times$  \\ 
\small$n\!=\!8100000,\ d\!=\!784$	\Cm				\Cm \textbf{\LibSVMMC}    \Cm 1.0   \Cm \textbf{1d 23h 12m 8s}  \Cm \textbf{6}$\times$ \\
\small$C\!=\!1000, \gamma\!=\!0.006$ \cite{bottou2006}
									\Cm	 \mclabel	\Cm \SPSVMMC    \Cm \red{1.4\std{?}}   \Cm \red{2h 37m 50s\std{?}} \Cm \red{115$\times$} \\

\hline
\end{tabular}
 
\caption{
Comparison of test error, training time, and speedup of kernelized SVM training methods.
The first column indicates dataset file size, number of instances, dimensionality, and SVM hyperparameters $C$ and $\gamma$ (with a citation for previously published values, otherwise derived by cross-validation using GTSVM).
Results for SP-SVM are the average of five runs with different randomly sampled candidate sets (see text for standard deviations).
Row background colors indicate implementation architecture: single-core (SC), \colorbox{blue1}{multi-core (MC)}, \colorbox{blue2}{GPU}.
\textcolor{red1}{Red font color} indicates poor test error results.
\textbf{Bold typeface} indicates the best timing results for each dataset and architecture.
Symbol $^\dagger$ indicates accuracy metric is $(1-$AUC$)\%$. 
Symbol \na{} indicates a data set/method pair that was unable to be run, as explained in the text.
}
 \label{table:speedup}
\end{center}
\end{table*}
 This section presents an empirical evaluation of several of the algorithms described in sections~\ref{section:explicit} and~\ref{section:implicit} on two modern parallel
architectures: multi-core CPUs (MC) and graphics
processing units (GPUs).  Running time and accuracy statistics on
seven datasets show the benefits and drawbacks
of the approaches included in our evaluation.

\paragraph{Hardware.}
Experiments are run on a 12-core machine with Intel Xeon X5650
processors at 2.67 GHz with hyperthreading enabled and 96 GB of RAM.
The attached NVIDIA Tesla C2075 graphics card contains 448 cores and 6
GB of global memory.

\paragraph{Methods evaluated.}
The \emph{single-threaded} CPU baseline method is LibSVM
\cite{chang2011libsvm}, a popular implementation of SMO, which we use as the baseline for classification accuracy.
On \emph{multi-cores} we evaluate a modified version of LibSVM which performs kernel computations in parallel with OpenMP\footnote{\urlT{http://www.csie.ntu.edu.tw/~cjlin/libsvm/faq.html}}. Further, we evaluate our implementation of \ourmethod{} 
in  \verb!MATLAB! with Intel MKL BLAS functions for matrix operations. 
For the \emph{GPU} settings, we compare two explicitly parallel GPU adaptations of dual decomposition:
{GPU SVM} \cite{catanzaro2008}, an adaptation of LibSVM for GPUs, and {GTSVM} \cite{cotter2011gtsvm}. We also include the implicitly parallel \verb!MATLAB! implementation of SP-SVM, linked against the appropriate libraries for GPU linear algebra computations. 
With the exception of SP-SVM, all implementations are written in \verb!C/C++! by the authors of the respective publications. 

\paragraph{Datasets.}
We evaluate all methods on several medium scale data sets, each involving classification tasks.
Medium scale datasets are chosen because parallel runtimes with small datasets tend to be dominated by overhead while large-scale datasets generally require an exorbitant amount of system memory. 
The datasets are as follows: Adult\footnote{\urlT{http://archive.ics.uci.edu/ml/datasets/Adult}}---an annual income prediction task (greater or less than $\$50$K) based on census data;
Covertype/Forest\footnote{\urlT{http://archive.ics.uci.edu/ml/datasets/Covertype}}---a tree cover prediction task based on geographical and climate features (predicting class 2 versus the rest);
KDDCup99\footnote{\urlT{http://kdd.ics.uci.edu/databases/kddcup99/kddcup99.html}}---a classification task for intrusion detection in network traffic;
MITFaces\footnote{\urlT{http://c2inet.sce.ntu.edu.sg/ivor/cvm.html}}---a face detection task from raw images (with accuracy presented in (1-AUC)$\%$ due to an extreme class imbalance);
Epsilon\footref{fn:pascal}---a synthetic classification task from the 2008 PASCAL Large Scale Learning Challenge;
FD\footnote{\urlT{http://largescale.ml.tu-berlin.de/instructions/}\label{fn:pascal}}---another face detection task (without heavy class imbalance); and
MNIST8M\footnote{\urlT{http://www.csie.ntu.edu.tw/~cjlin/libsvmtools/datasets/multiclass.html}}---a multiclass handwritten digit recognition task based on label invariant transformations of images from the MNIST data set. We use the one-versus-one classifier approach to multi-class classification, as also adopted by LibSVM~\cite{chang2011libsvm}.

Features for the datasets Adult, Covertype/Forest, KDDCup99, MITFaces, and
MNIST8M are scaled to $[0,1]$ before training.  In addition, we
subsample two of the largest data sets, Epsilon and FD, uniformly at
random from $400,000$ to $160,000$ and $5,469,800$ to $200,000$
respectively for two reasons.  First, single core algorithms
require prohibitively long training times on the full sets.  Second,
on GPUs, if the data does not fit into GPU memory the running time is
dominated by memory transfer, which is not the focus of this study.

\textbf{Hyper parameters.}
The left column of Table~\ref{table:speedup} provides details of the size
and dimensionality of each data set.  In addition, it also indicates
the regularization parameter $C$ and inverse Gaussian kernel
width $\gamma$ used throughout the experiments.  These parameters are
derived from cited works for most datasets, as indicated in the table.
For Epsilon and FD, a thorough cross-validation grid search
was conducted using GTSVM as it is an
exact implementation and tends to behave identically to LibSVM in
terms of hyper parameters but does not have the large time requirement
of cross validating with LibSVM. This approach does a slight disservice to
SP-SVM, however it may be viewed as a fair compromise as LibSVM is the gold standard and our main focus is the speedup.
Throughout all experiments with SP-SVM we set the stopping criterion to $\epsilon\!=\!5\!\times\! 10^{-6}$. 

\textbf{Evaluation.} 
Table \ref{table:speedup} shows test error, training time, and
speedup versus single-core LibSVM for all methods on each of the seven
data sets.  The training times omit both loading data from disk and
computing test predictions for all methods.  As MNIST8M is multi-class, the times reported are the accumulative
time for each one-versus-one classifier trained individually.\footnote{Shared memory computers, such as multi-core CPUs and GPUs, are arguably less suited for this kind of multi-class classification, since  
one-versus-one classifiers are ``embarrassingly parallel'' for
problems with many classes and can be solved on (cheaper) distributed
memory machines (clusters) with near-perfect speedup.}

Since SP-SVM deploys a heuristic based on random sampling of basis
vectors, we computed five runs for each setting and report the average
runtime and test error.  Standard deviations on SP-SVM test error are
less than $0.001$ for all datasets except for the multicore
implementation on KDDCup99 ($0.0023$).
Similarly, standard deviations for SP-SVM training time are on the
order of seconds for each run. (For increased readability, we omit
them from the table.)  

Not all algorithms converge on all data sets. GTSVM is the only GPU method that runs on
KDDCup99 (which is $90\%$ sparse). GPU SVM and SP-SVM both store the inputs in dense format on the GPU, which exceed its memory. The dense MNIST8M data is too large for all
GPU algorithms.\footnote{As GPU memory sizes grow, this limitation will become less important. In addition, GPUs and CPUs might eventually converge on using a single memory space. For sparse data sets one might also consider special purpose libraries, such as CUSPARSE (\url{https://developer.nvidia.com/cusparse}), for the kernel computation.}
Also, LibSVM with OpenMP failed to converge on Epsilon in less time than single-core LibSVM.

\textbf{Accuracy.} For most datasets and methods, test errors are remarkably consistent,
even between exact and approximate methods. 
However there are a few notable exceptions,
highlighted in red in Table \ref{table:speedup}. 
GTSVM fails on Covertype/Forest and we hypothesize that this anomaly may be
due a floating point precision error as the method converges when run on smaller subsets of
the training data. On KDDCup99 GTSVM obtains an error
rate of $19.9\%$, which is not significantly better than a constant predicting the most
common class (no GPU method in our evaluation could successfully learn from this data). SP-SVM performs slightly worse on KDDCup99 ($7.9\%$ vs. $7.4\%$) and noticeably worse on MITFaces ($7.4\%$ 1-AUC vs. $5.6\%$) and MNIST8M ($1.4\%$
vs. $1.0\%$). The approximation error may be more pronounced on MITFaces due to the large class imbalance (a few additional false positives have a strong effect on the final area under the curve) and also for MNIST8M, where  the approximation error is being
aggregated across the many (45) one-versus-one classifiers. 

\textbf{Speedup.}
The most basic method of speedup is LibSVM
on multicores. This involves a trivial change directly to the source of LibSVM,
allowing for the use of OpenMP parallel for-loops in kernel
computations.  Because kernel computations account for such a
significant portion of LibSVM's runtime, this baseline improvement
results in a $5-8\times$ speedup on twelve cores. 

GPU SVM achieves  $20-40\times$ speedups over
single-core LibSVM by performing kernel computations and KKT condition updates directly on the GPU.  GTSVM achieves the largest speedups among the dual decomposition methods, by also increasing the working set size to
$16$ (compared to $2$ used by LibSVM and GPU SVM), leading to
$2.5-6.5\times$ speedup over GPU SVM, and $2-250\times$ speedup over
LibSVM.  This highlights the correlation between speedup and the amount of handcrafted parallelism that is included in the algorithm design for the explicit parallel approaches. 

 In comparison to single
core LibSVM, SP-SVM achieves $13\times$ to $193\times$ speedup on
multi-cores, and $17\times$ up to $601\times$ speedup on GPUs. On
both architectures, the speedup of SP-SVM tends to increase with data
set size, which reflects the increasing time spent inside parallelized
library operations. The smallest speedup for both architectures is on
the smallest data set, Adult---however, by a mere $11s$ or $15s$
compared to the fastest algorithm (GTSVM).
It is surprising just how effective the parallelism derived from the dense linear algebra in SP-SVM proves to be on both architectures.  SP-SVM is particularly effective on GPUs where it
outperforms all other GPU methods by $1.5\times$ to $5\times$ on all
but Adult, and achieves a $1.3-4.3\times$ speedup over multi-core
SP-SVM. However even on multi-cores, SP-SVM  outperforms GPU SVM and GTSVM significantly on  MITFaces and FD.
SP-SVM requires only $11$ minutes on average across all binary
classification datasets, compared to the several hours often required
by LibSVM.

\section{Discussion}\label{section:discussion}
One trend clearly follows from our study: massive speedups are possible when the parallelism of modern hardware is leveraged. 
Although explicit parallelization is by far the most dominant approach to SVM parallelization, our results demonstrate that implicit parallelization can be more efficient and deserves some attention. We believe that the community can benefit from our findings in two ways: first, practitioners will obtain an easy to use implementation of SP-SVM with unprecedented training speed that can readily be used on most platforms with BLAS compatible libraries; second, researchers working on parallel machine learning algorithms may reconsider spending days in agony on C/C++ programming of parallel code and may instead focus on  relaxing or reformulating their algorithm to rely more heavily on dense linear Algebra routines. 
 
One downside of implicit parallelization for SVMs is that the exact reformulations are too memory intensive, and SP-SVM enforces a reduced basis vector set. We show in our results that in practice this effect might be small, however it will be interesting to see if there are exact formulations that avoid such restrictions. 
We predict that relaxations into implicit parallelization may become increasingly important as multi-cores and GPUs establish themselves as the common computing platforms, similar  to relaxing optimization problems into convexity, as has been common practice for years.

\subsubsection*{Acknowledgments}
ST and KQW are supported by NSF grants 1149882 and 1137211, ST and KA are supported by NSF grants 1150036 and 1218017, and ST is supported by an NVIDIA Graduate Fellowship. The authors thank Olivier Chapelle for sharing his Matlab implementation of SP-SVM, and Gabriel Hope, Nicholas Kolkin and Jack Hessel for writing the C++ version of WU-SVM.

%
\bibfontsize
\bibliographystyle{abbrv}
\bibliography{gpusvm}  

\end{document}